\pdfoutput=1
\documentclass{IEEEaccess}
\usepackage{cite}  %FGF: commented this to make biblatex work
\usepackage{amsmath,amssymb,amsfonts}
\usepackage{algorithmic}
\usepackage{graphicx}
\usepackage{textcomp}
\usepackage{graphics}
\usepackage{booktabs}
\usepackage{xurl}
\usepackage{color, soul}
\usepackage{array}
\usepackage{tabularx}
\usepackage{multirow}
\usepackage{subcaption} % put in preamble

\def\BibTeX{{\rm B\kern-.05em{\sc i\kern-.025em b}\kern-.08em
    T\kern-.1667em\lower.7ex\hbox{E}\kern-.125emX}}

\makeatletter
\def\endthebibliography{%
  \def\@noitemerr{\@latex@warning{Empty `thebibliography' environment}}%
  \endlist
}
\makeatother

\begin{document}
\history{Date of publication xxxx 00, 0000, date of current version xxxx 00, 0000.}
\doi{}

\title{A Lightweight 3D-CNN for Event-Based Human Action Recognition with Privacy-Preserving Potential}
\author{
\uppercase{Mehdi Sefidgar Dilmaghani\authorrefmark{1}, Francis Fowley
\authorrefmark{1}, and Peter Corcoran}\authorrefmark{1},
\IEEEmembership{Fellow, IEEE}}
\address[1]{Department of Electronic Engineering, College of Science and Engineering, University of Galway, Galway, H91 TK33 Ireland}

% ************Is this PRIVI-SENS?***********
\tfootnote{``This work was supported in part by the Irish Research Council (Grant No. IRCLA/2023/1992).''}

\markboth
{Author \headeretal: Preparation of Papers for IEEE TRANSACTIONS and JOURNALS}
{Author \headeretal: Preparation of Papers for IEEE TRANSACTIONS and JOURNALS}

\corresp{Corresponding author: Sefidgar Dilmaghani, M (e-mail: 
mehdi.sefidgardilmaghani@universityofgalway.ie).}

\begin{abstract}
This paper presents a lightweight three-dimensional convolutional neural network (3D-CNN) for human activity recognition (HAR) using event-based vision data. Privacy preservation is a key challenge in human monitoring systems, as conventional frame-based cameras capture identifiable personal information. In contrast, event cameras record only changes in pixel intensity, providing an inherently privacy-preserving sensing modality. The proposed network effectively models both spatial and temporal dynamics while maintaining a compact design suitable for edge deployment. To address class imbalance and enhance generalization, focal loss with class reweighting and targeted data augmentation strategies are employed. The model is trained and evaluated on a composite dataset derived from the Toyota Smart Home and ETRI datasets. Experimental results demonstrate an F1-score of 0.9415 and an overall accuracy of 94.17\%, outperforming benchmark 3D-CNN architectures such as C3D, ResNet3D, and MC3\_18 by up to 3\%. These results highlight the potential of event-based deep learning for developing accurate, efficient, and privacy-aware human action recognition systems suitable for real-world edge applications.
\end{abstract}


\titlepgskip=-15pt

\maketitle

%%%%%%%%%%%%%%%%%%%%%%%%%%%%%%%%%%%%
\section{Introduction}\label{Introduction}
% Background
\PARstart{H}{uman} Action Recognition (HAR), using cameras and other sensors such as radar, plays a critical role in the care of elderly, sick, and disabled individuals by reducing the need for constant monitoring by human caregivers. HAR also has a wide range of other applications, including intelligent surveillance, human–computer interaction, and smart environments \cite{M8}. Recognizing an action requires not only understanding what is happening in each frame at a given moment but also how it changes over time to make an accurate prediction. As an example, pouring hot water into a cup could be part of making tea, instant coffee, or another drink, so the action needs to be observed over time to determine the exact outcome. Other challenges also exist in making HAR practical for real-world applications. These include limitations in the sensors, and in the algorithms \cite{F34}.

% Challenges in Human Action Recognition
Regarding sensors, most current HAR algorithms rely on frame-based cameras, which must continuously capture frames to monitor human activities. These systems not only need to store the frames but also process them using preprocessing and deep learning algorithms, requiring substantial memory and computational resources. A more critical limitation is that such cameras record personal information in private spaces, raising privacy concerns and often conflicting with regulations and user consent \cite{M9}. 

On the algorithm side, significant challenges also remain. Networks that achieve high accuracy have complex architectures and require considerable computational and memory resources. Deploying such algorithms on edge devices is even more challenging, and their design must be highly efficient and optimized. Consequently, there is a growing need for lightweight, high accuracy, and privacy-preserving HAR solutions that can operate efficiently without compromising recognition performance \cite{M10}.

% Event Cameras and Their Benefits
Recent advances in deep learning have shown great potential in addressing these challenges. Deep neural networks have been developed with a strong ability to learn both spatial and temporal patterns from large datasets \cite{M4}. New algorithms have also improved the accuracy of HAR systems, making it possible to build models that can understand, predict, and respond to human behavior in real time. In addition, significant research has focused on designing compact neural networks suitable for deployment on edge devices \cite{M11}.

To address privacy concerns, event cameras—also known as neuromorphic or dynamic vision sensors—offer a promising alternative to conventional frame-based cameras. In contrast to standard cameras, which capture full intensity frames at fixed intervals, event cameras record only changes in pixel intensity, producing a stream of events that represent motion. Because they do not capture detailed texture or color information, event cameras inherently preserve privacy, making them particularly suitable for sensitive HAR applications \cite{M4}.

% Limitations of Existing Methods
Despite the potential of novel AI algorithms and event cameras, research on combining these two for HAR applications remains relatively limited. Most existing HAR methods have been developed for frame-based video data using convolutional neural networks (CNNs). While these models have demonstrated strong performance, they face limitations beyond those mentioned above; in particular, CNNs are not well suited to capturing temporal information \cite{M12}.

The few existing approaches that leverage event data often rely on complex architectures, large model sizes, or extensive preprocessing pipelines to convert asynchronous events into other representations. These requirements reduce the advantages of event cameras and limit their applicability in real-time or embedded systems. Consequently, there is a growing need for simpler, yet effective, deep learning models specifically designed for event-based data \cite{M13}.

% Contributions of This Work
To address these challenges, we have designed and trained a three-dimensional convolutional neural network (3D-CNN) to analyze the output of event cameras. 3D-CNNs have the same capability as conventional CNNs in capturing spatial information, while also extracting temporal features from consecutive frames. Using event cameras instead of RGB cameras represents a step toward privacy-preserving HAR. Although the event stream is collected in 2D matrices to be fed into the 3D-CNN, this approach is slightly more complex than traditional frame-based methods that directly generate full frames.

The goal of this paper is to propose an algorithm that analyzes event camera data with a model that is accurate, compact, and privacy-preserving. To meet these requirements, our contributions are:

\begin{itemize}
    \item Proposing a 3D-CNN specifically designed for HAR that captures both spatial and temporal information,
    \item Training the algorithm using event data to support privacy preservation,
    \item Achieving efficient and robust recognition with a lightweight network without requiring large computational resources.
\end{itemize}

% Paper Organization
The rest of the paper is organized as follows. Section II reviews related work on HAR. Section III presents the proposed methodology in detail. Section IV describes the experimental results and discusses the findings. Finally, Section V concludes the paper and outlines directions for future research.
%%%%%%%%%%%%%%%%%%%%%%%%%%%%%%%%%%%%%%%%
\section{literature review}\label{literature review}
Human action recognition has been an important problem in computer vision, driven by applications in surveillance, healthcare, and human–computer interaction. Numerous approaches have been proposed over the past decades to address this challenge. However, several issues remain unresolved and continue to be active areas of research, particularly privacy concerns \cite{M6} and the difficulty of accurately recognizing visually similar activities \cite{M7}. The following section provides a review of related work in this domain.

\subsection{Traditional Image and Video Processing Methods for HAR}
Prior to the vast developments in deep learning based methods, HAR was based on hand-crafted spatio-temporal features, bag-of-features, and trajectory representations. Space-time interest points (STIPs) detect localized spatio-temporal saliency and were foundational for early video descriptors \cite{F1}.
Dollár et al. proposed sparse spatio-temporal cuboid features and demonstrated robust behavior/gesture recognition with these local descriptors \cite{F2}.
Dense trajectory methods and their improvement via camera-motion compensation became a dominant representation for action recognition because they encode long-term motion patterns robustly \cite{F3}.
These methods report high performance on curated benchmarks of their era (e.g., improved dense trajectories reach competitive accuracies on UCF-101 / HMDB benchmarks) but remain sensitive to viewpoint, illumination and require manual feature design. The strengths and limitations of these approaches are summarized in the comprehensive survey proposed in \cite{F4}.

\subsection{Deep Learning with Frame-based Cameras for HAR}
Deep learning moved HAR away from engineered features to learned spatio-temporal representations. It introduced architectures such as Two-Stream CNNs \cite{F5}, large-scale video CNNs \cite{F6}, Temporal Segment Networks (TSN) \cite{F7}, and Inflated 3D ConvNets (I3D) \cite{F8}, which demonstrated strong performance but at significant computational cost.

The two-stream fusion of an RGB (spatial) net and an optical-flow (temporal) net achieved strong gains on standard benchmarks. The fused two-stream model reported 88.0\% mean accuracy on UCF-101 and 59.4\% on HMDB-51 when using SVM fusion; temporal stream alone was 83.7\% on UCF-101 in the original experiments. These results established that explicit modeling of motion via optical flow substantially boosts accuracy \cite{F5}. TSN introduced sparse but global temporal sampling and aggregation. The paper reports 94.9\% accuracy on UCF-101 and 71.0\% on HMDB-51. TSN also offers efficient variants that trade off accuracy and speed \cite{F7}. Work on large-scale datasets and architectures established design patterns and the need for scale \cite{F6}. Inflated 3D ConvNets (I3D) extended 2D filters to 3D and, trained on the Kinetics dataset, set new state-of-the-art baselines for learned spatio-temporal modeling \cite{F8}.

C3D, a 3D ConvNet trained end-to-end on video volumes,  reported 85.2\% on UCF-101 which was one of the best prior 3D convolutional results at the time and strong gains on Sports-1M. These results demonstrated the value of spatio-temporal convolutions \cite{M1}. R(2+1)D is an architecture that factorizes 3D convs into separate spatial and temporal convolutions. Its variants reached state-of-the-art levels on large video datasets. Reported video level accuracies include 73.3\% on Sports-1M and strong Kinetics numbers in the original study \cite{F10}. SlowFast introduces dual pathways (Slow and Fast). Depending on instantiation, SlowFast variants achieve 75–80\% accuracy on Kinetics-400. The work also reports FLOPs, inference-view tradeoffs, and shows consistent gains over single-path baselines \cite{F11}. Temporal Shift Module (TSM) is a lightweight temporal module enabling 2D backbones to capture temporal information with low overhead. Reported Kinetics performance for TSM-ResNet50 is 74.1\% accuracy. A mobile TSM variant with MobileNet-V2 achieved 69.5\% accuracy on Kinetics, demonstrating feasibility for edge devices \cite{F12}. Direct numeric comparisons must be read carefully. That is because evaluation protocols including number of clips, multi-crop testing, pretraining, and dataset splits differ for different projects. Transformer token-based and well-tuned 3D convs often lead on Kinetics scale benchmarks, while two-stream and TSN remain strong and more efficient on some tasks \cite{F5, F7}.

Although the mentioned architectures achieve high accuracy in certain cases, they do not address privacy concerns. Since all of them require frames containing personal information for training and inference, they are not suitable for real-world applications, especially under regulations such as the GDPR.

\subsection{Privacy-Preserving HAR}
The aforementioned HAR methods require frames containing personal information for correct functioning, which raises privacy concerns. Alternative solutions include RF-based systems such as WiSee \cite{F13}, TARF \cite{F14}, WiFall \cite{F15}, and the radar-based TinyRadarNN \cite{F16}, which achieve competitive accuracy without visual recording.

WiSee \cite{F13} was an early system that used WiFi Doppler signals for whole-home gesture recognition, highlighting the through-wall capability and privacy benefits of RF sensing. The paper reports an average accuracy of 94\% for nine gestures in both in-room and through-wall tests, with reduced performance when multiple people are present. Commercial mmWave (60 GHz / 77 GHz) radars and recent research prototypes demonstrate real-time action and gesture recognition using micro-Doppler features and lightweight classifiers, making radar a practical and privacy-preserving choice for in-home HAR \cite{F16, F17, F18}.

TinyRadarNN \cite{F16} is a compact, low-power, radar-based gesture classifier with only 46k parameters and a memory footprint of 92 KB. On its 11-gesture dataset, it achieves 86.6\% accuracy in multi-user cross-subject tests and 92.4\% in single-user experiments, while consuming just 21 mW during inference. This strong performance underscores radar’s potential for embedded HAR \cite{F16, F19, F21}.

Comprehensive reviews of radar, WiFi, and RFID-based HAR approaches, their machine learning pipelines, and challenges such as domain adaptation and multimodal fusion are provided in \cite{F14, F15}. Although RF and radar methods demonstrate strong accuracy on constrained gesture sets and operate with low power consumption, they are often evaluated on different gesture vocabularies and under controlled conditions, complicating cross-method comparisons \cite{F20}.

% More event works required here
\subsection{Event-based Vision for HAR} 
Event cameras, or neuromorphic sensors, generate a stream of events rather than standard image frames. They record asynchronous brightness changes at the pixel level with microsecond latency and high dynamic range. These properties make them attractive for low-power and privacy-sensitive applications, including HAR\cite{M4}. Several systems \cite{F22}, benchmarks \cite{F23}, and datasets \cite{F24} have been developed for this purpose, demonstrating the potential of event-based sensing for efficient HAR.

Early event-based gesture and activity recognition systems converted event streams into 2D matrix representations for conventional processing \cite{M5}. More recent benchmarks and dedicated action-recognition datasets demonstrate growing interest in event-based HAR, introducing evaluation protocols that emphasize temporal ordering rather than static event accumulation \cite{F24}.

\subsection{3D CNNs with Event Cameras for HAR}
Algorithms that use both spatial and temporal information are effective for tasks such as HAR. In this area, spatio-temporal deep learning with event-based data has become an important approach. These methods often use 3D-CNNs \cite{M1}, voxel-based models \cite{F25}, asynchronous networks \cite{F26}, and learned event representations \cite{F27}. By combining spatial features with changes over time, these techniques can better understand motion and activities in complex environments.

A common approach for applying deep video models to event-based data is to convert the asynchronous event stream into dense spatio-temporal representations, such as voxel grids, event spike tensors, or time surfaces. These representations are then processed using 2D/3D or 3D CNN backbones, such as C3D and I3D. While C3D and related 3D CNN architectures effectively capture spatio-temporal features, they are computationally expensive and memory-intensive, which poses challenges for deployment on embedded or edge devices \cite{M1, F8}.

Gehrig et al. formalized a set of grid-based event representations, including event frames, voxel grids, and event spike tensors, which serve as inputs to conventional CNNs. Their work highlighted the trade-off between preserving temporal resolution and managing computational cost \cite{F27}. Recent event-specific architectures explore voxel, graph, and transformer-based representations, e.g., voxel-graph CNNs, and volumetric transformers, to exploit event sparsity. However, models employing full 3D convolutional stacks still have high parameter counts and computational demands, motivating the development of sparsity-aware, quantized, or lightweight alternatives \cite{F4, F25, F27}.

\subsection{Lightweight Deep Learning Models for Vision Tasks}
Efforts to reduce computational cost have focused on developing compact, high accuracy models for resource constrained platforms, leading to lightweight HAR architectures suitable for edge deployment. Representative model families include MobileNet, which employs depthwise separable convolutions \cite{F29}; MobileNetV2, which introduces inverted residuals and linear bottlenecks \cite{F30}; ShuffleNet, which utilizes channel shuffling and group convolutions \cite{F31}; and EfficientNet, which applies compound scaling based on neural architecture search \cite{F32}. These architectures achieve favorable accuracy–latency trade offs, making them well suited for efficient on-device inference.

These models have been adapted for tasks such as detection and segmentation, and extended to video through 2D backbones with temporal fusion or efficient 3D temporal modules. When integrated with sparse event encodings or lightweight temporal components, such architectures form strong foundations for efficient HAR systems, including event based pipelines. Mobile oriented backbones when combined with temporal modules such as TSM, provide favorable accuracy efficiency trade offs. For example, an online TSM variant using a MobileNetV2 backbone achieved approximately 69.5\% accuracy on the Kinetics dataset while remaining practical for edge deployment. These results highlight the potential of mobile backbones for event or radar based low power HAR applications \cite{F33}.

\subsection{Summary of HAR Models}
Table \ref{tab:models performances} summarizes the key HAR models reviewed. It lists the model type, the data modality used, and the reported performance metrics. The table illustrates the progression from traditional handcrafted features to deep learning approaches, including 3D CNNs, two-stream networks, transformer-based models, and lightweight architectures suitable for resource-constrained environments. It also highlights computational aspects, such as model size and efficiency, for each approach.

\begin{table}[ht]
\centering
\caption{Reported performance of representative HAR models}
\begin{tabular}{|l|c|c|c|}
%{|p{35pt}|p{20pt}|p{50pt}|p{20pt}|p{25pt}|p{45pt}|} 
\hline
\textbf{Model} & \textbf{Dataset} & \textbf{Accuracy / Top-1} & \textbf{Citation} \\\hline
    Two-Stream & UCF-101 & 88.0\% & \cite{F5} \\
               & HMDB-51 & 59.4\% &  \\ \hline
    C3D        & UCF-101 & 85.2\% & \cite{M1} \\\hline
    R(2+1)D    & Sports-1M & 73.3\% & \cite{M3} \\\hline
    TSN        & UCF-101 & 94.9\% & \cite{F7} \\
               & HMDB-51 & 71.0\% &  \\\hline
    I3D        & Kinetics-400 & 72\% & \cite{F8} \\\hline
    SlowFast   & Kinetics-400 & 79.8\% & \cite{F11} \\\hline
    TSM-Res50  & Kinetics-400 & 74.1\% & \cite{F12} \\\hline
    TSM-Mobilenet & Kinetics-400 & 69.5\% & \cite{F12} \\\hline
    WiSee      & 9 Gestures & 94\% & \cite{F13} \\\hline
    TinyRadarNN& 11 Gestures & 86--92\% & \cite{F21} \\\hline
    Swin-T     & DailyDVS-200 & 48.1\% & \cite{F19} \\\hline
    Spikformer & DailyDVS-200 & 36.9\% & \cite{F19} \\\hline
    EVMamba    & DailyDVS-200 & +8.8\% vs. TSM & \cite{F20} \\\hline
    
\end{tabular}
\label{tab:models performances} 
\end{table}

\subsection{Summary of Gaps in Literature}
Despite the considerable achievements in the field of HAR, especially following recent developments in AI algorithms, the review presented in this section indicates that several areas still require further attention and remain far from the ideal state. The following fields, in particular, need more focus:

\textbf{Privacy–accuracy trade-offs:} Radar and WiFi based methods protect privacy better than camera systems but still are less accurate in recognizing detailed actions and functioning across different scenes. Combining multiple sensing methods is still at an early stage \cite{F17, F14}. These approaches can reach high accuracy for small, fixed gesture sets (e.g., WiSee: 94\%; TinyRadarNN: 86–92\%), but fair comparison with camera based methods is difficult because they use different datasets and problem setups. The main challenge is to match the detail of camera based recognition while keeping user privacy \cite{F34}.

\textbf{Efficient spatio-temporal modeling for events:} Voxelization and 3D CNNs preserve temporal details but result in large models that are costly for edge deployment. Sparsity-aware, asynchronous, and spiking approaches show promise but still require thorough comparison with compact ANN-based alternatives \cite{F27, F25}.

\textbf{Compute vs. temporal fidelity:} 3D CNNs reach high accuracy on large video datasets such as Kinetics, however require heavy computational and memory resources. Lighter models such as TSM or MobileNet cut these costs with reasonable drops in accuracy. This trade-off is still not well studied for event based and RF data \cite{F35}. While efficient image networks are widely used, their utilization for temporal modeling in HAR is limited. Adapting architectures such as MobileNet for event or RF inputs is challenging and remains an open area for research \cite{F29}.

\textbf{Event-based dataset scale and model gap:} Most event HAR datasets are small or can be analyzed using simple, non-temporal methods. There is a clear shortage of large, realistic datasets that require temporal modeling and generalization across subjects and scenes \cite{F36, F37}. Recent results on the DailyDVS-200 dataset show that transformer and token-based models currently perform best on large scale event HAR, but top-1 accuracy remains below 50\% for cross-subject evaluation. While models such as EVMamba show promising improvements, they still lag behind dense frame based methods on simpler tasks. These results highlight the requirement for better event specific architectures and larger cross domain benchmarks \cite{F38}.

%%%%%%%%%%%%%%%%%%%%%%%%%%%%%%%%%%%%%%%%
\section{Methodology}\label{Methodology}
This section explains the details of the algorithm that’s proposed to recognize the human activities. The algorithm is a 3D-CNN that extracts the information from the raw stream of events generated by a neuromorphic sensor. The reason behind using the 3D-CNN rather than other algorithms is its strength in analyzing both spatial and temporal information, simultaneously.

The designed algorithm is a lightweight network which leads to faster training process and easier implementation on the edge devices. Beyond the advantages of the proposed network, output data of the event cameras offer inherent privacy-preserving potential. Although in the proposed method, the raw events still need to be collected into 2D matrices, which resemble frames, accessing personal information remains more complex than with frame-based cameras. The proposed pipeline consists of preprocessing input data, feeding them into a lightweight yet expressive 3D-CNN model, and training the network. A self-attention is incorporated besides the 3D convolutional layers to enhance feature representation of the network if required. Other strategies in the algorithm design include utilization of focal loss and class reweighting. Th details of the pipeline are explained in this section.

\subsection{Event Data Representation} 
The main dataset used to train the network consists of RGB videos of varying lengths. These videos are first converted into raw event data, which are then accumulated into 2D matrices at a rate of 30 fps. Consequently, the total number of 2D matrices—hereafter referred to as event frames—varies across videos. However, in 3D-CNNs for proper processing, the number of frames must be consistent across batches of videos. To solve this problem, a consistent number of event frames per video are downsampled uniformly, and the intermediate frames are discarded. This consistent rate is set fixed at 10 frames per video as this rate achieved a good balance between accuracy and GPU load. Moreover, as will be shown later, lower rates resulted in a considerable drop in accuracy, whereas higher rates achieved no improvement. Detailed information about the dataset preparation is as follows.

\subsubsection{Training Dataset}
The reason for using an RGB dataset rather than an event-based one is the limited availability of suitable event-based datasets for HAR applications. Even among non-event datasets, most are constrained by data privacy concerns, minimal scene diversity, the requirement for expensive sensor equipment, and an insufficient number of participants \cite{F39}. 

Natural realism is a major challenge in building effective HAR datasets. Scripted scenes performed by actors often lack spontaneity and natural occlusions. As noted by Htun et al. \cite{F39}, datasets designed mainly for object detection but without sufficient human–object interactions are “context-biased,” which reduces the performance of HAR models. In addition, background environments in home-like settings are difficult to modify, leading to limited variation in object placement, such as furniture or appliances. Virtual-reality simulations can help address this issue, but realism remains limited. Finally, most reviewed datasets show little correspondence between activity classes and the instruments used to perform them.

The Toyota Smart Home  (TSH) trimmed dataset and the ETRI dataset were selected as the most suitable sources for compiling the training data for the 3D-CNN model. Both datasets contain a large number of video samples, cover a wide range of annotated activities, and include challenges such as occlusions and lighting variations. Common subclasses were identified and extracted from both datasets, and subsequently merged into six classes for training, as summarized in Table \ref{tab:training_dataset}. To create a balanced training dataset, ETRI videos were segmented into shorter clips: videos of 5–10 seconds were split in half, those of 10–15 seconds into thirds, and those longer than 20 seconds into four equal segments. All videos were resized to a frame resolution of 640 × 480 pixels. To address differences in dataset size, 720 video samples were selected from each ETRI class, and 280 samples were randomly drawn from each TSH class, resulting in a balanced dataset of 1,000 samples per HAR class.

\newcolumntype{C}[1]{>{\centering\let\newline\\\arraybackslash\hspace{0pt}}m{#1}}

\begin{table}[htp]
\caption{Balanced training dataset compiled from ETRI and TSH sources.}
\centering
\begin{tabular}{|C{1.8cm}|C{1.5cm}|C{1.5cm}|C{1.5cm}|}
\hline
\textbf{Class} & \textbf{Samples from TSH} & \textbf{Samples from ETRI} & \textbf{Total} \\ \hline
Cooking       & 280 & 720 & 1,000 \\ \hline
Drinking      & 280 & 720 & 1,000 \\ \hline
Eating        & 280 & 720 & 1,000 \\ \hline
Getting Up    & 280 & 720 & 1,000 \\ \hline
Sitting Down  & 280 & 720 & 1,000 \\ \hline
Washing Up    & 280 & 720 & 1,000 \\
\hline
\end{tabular}
\label{tab:training_dataset}
\end{table}

\subsubsection{Data Augmentation}
The data augmentation strategy is targeted and designed to improve generalization while addressing class imbalance. While standard resizing, padding, and normalization is used for all of the classes, for underrepresented classes which include "Eating" and "Washing Up", random horizontal flip, rotation, affine transform, and Gaussian blur is used.

\subsection{Network Architecture}
The proposed lightweight 3D-CNN uses a series of 3D convolutional layers each being followed by a batch normalization and pooling layer. An average pooling and a fully connected layer for classification are the final layers.

\subsubsection{3D Convolution Layers}
The network includes five sequential 3D convolutional blocks with increasing channel sizes: 1, 16, 32, 64, 128, and 256. Each convolution is followed by BatchNorm3d, ReLU activation, and MaxPool3d to downsample spatial dimensions while preserving the temporal dimension (kernel\_size= (1,2,2)). This architecture allows the model to learn both spatial and temporal features concurrently.

\subsubsection{Classification Head}
The final classifier consists of a dropout layer for regularization, a global average pooling for dimensionality reduction, and a fully connected layer that maps the pooled features to class probabilities. This minimal design ensures a balance between efficiency and performance.

\subsubsection{Feature Extraction}
The 3D convolution layers operate over the frame sequence to extract spatial-temporal features. The final average pooling layer reduces the feature map size to provide a compact representation of the input video for the classification head. 

\subsubsection{Loss Function and Optimization}
The model is trained using Focal Loss, which addresses class imbalance and emphasizes on the examples which are more difficult to classify. The focal loss is defined as:

\begin{equation}
\text{FL}(p_t) = -\alpha_t (1 - p_t)^\gamma \log(p_t)
\label{eq:focal_loss}
\end{equation}

where \(p_t\) is the predicted probability of the true class, \(\alpha_t\) is a weighting factor inversely proportional to class frequencies to handle class imbalance, and \(\gamma = 2.0\) is a focusing parameter that down weights easy examples. This combination allows the network to focus on videos which are harder to classify.  

The AdamW optimizer is used for training, chosen for its adaptive learning rate and decoupled weight decay, which improves generalization. Early stopping is applied to prevent overfitting during training.  

\subsubsection{Model Efficiency and Design Choices}
The main concern during the development process was to keep the model lightweight with limited parameters to ensure faster training and inference, and easier implementation on edge devices. The final average pooling removes the requirement for large fully connected layers. Self-attention module is optional and can be toggled to balance complexity with performance. The operation can be conducted efficiently on limited GPU resources (e.g., batch size 32 with pinned memory and num\_workers=8).

\subsection{Training Strategy}
During the training procedure the AdamW optimizer learning rate is fixed to 0.0009 and weight\_decay is 1e-4. Early stopping is designed with a patience of 20 epochs based on F1 score improvements. Best-performing model is saved and later evaluated on a held-out test set. Training and validation losses are logged and saved as plots and will be presented in the results and discussion section.

\subsection{Privacy-Preserving Characteristics}
By collecting the event-based data into low-resolution grayscale 2D matrices, the model inherently provides a degree of privacy as the event cameras do not capture identifiable details. This makes the approach more suitable for privacy-sensitive applications such as healthcare or surveillance.

%%%%%%%%%%%%%%%%%%%%%%%%%%%%%%%%%%%%%%%%%%%%%%%%%%%%%%%%%%%%%%
\section{Results and Discussion} \label{Results and Discussion}
In this section, we first present and analyze the results of training, validation, and testing of the proposed structure on the prepared dataset. Next, we train three widely used 3D-CNN architectures to enable a fair comparison between the performance of our proposed structure and state-of-the-art models.

\subsection{Description of Experiment}
The model is trained on grayscale sequences of 2D matrices extracted from stream of event data. Each sequence (video) comprises 10 temporally spaced frames, representing a short-duration action segment. A lightweight 3D CNN architecture is employed, incorporating optional self-attention, with focal loss to counteract class imbalance. Early stopping is applied to avoid overfitting. Random shuffling ensures fair train-validation-test splits.

\subsection{Description of Dataset}
All training, validation, and testing are conducted on a custom event dataset comprising short videos of human activities. The dataset contains six classes: cooking, washing dishes, sitting down, getting up, eating, and drinking. These classes are selected in pairs with similar characteristics to ensure a more challenging classification task. Each class includes 1,000 sequences, with each sequence represented as a folder containing image frames. Data augmentation techniques, as described in the methodology section, are applied to increase variability. Each input sequence consists of 10 grayscale frames, resized and padded to a resolution of 128×128.

\subsection{Evaluation Metrics}
Standard multi-class classification metrics are employed for evaluation. The following metrics are used:

\begin{itemize}
    \item Focal Loss – Applied during both training and validation. The details of this loss are explained in the methodology section.
    \item Accuracy – Measures the overall proportion of correctly predicted samples.
    \item F1-Score – The harmonic mean of precision and recall, computed with weighted averaging to account for potential class imbalance.
    \item Confusion Matrix – Provides a visualization of misclassifications across classes.
    \item Training Time – Reported to enable fair comparison between the proposed method and state-of-the-art networks.
\end{itemize}

\subsection{Hardware and Training Environment}
All experiments are conducted on a workstation equipped with an NVIDIA GeForce RTX 3090 Ti GPU with 24 GB of VRAM, an Intel Core i7-7700K CPU @ 4.20 GHz (4 cores, 8 threads), and 64 GB of system memory. The system runs Ubuntu 22.04.5 LTS (kernel 6.8.0-79-generic) with NVIDIA driver version 570.169. The software environment is based on Python 3.10.12 and PyTorch 2.5.1+cu121, which is compiled with CUDA 12.1 and linked against cuDNN 9.0.1. The batch Size is 32 for all data loaders, epochs are set to 1000 with early stopping with patience of 100 epochs. Learning Rate is 0.0009 and AdamW is used as optimizer.

\subsection{Quantitative Results}
The proposed model achieved a test accuracy of 94.17\% and a test F1-score of 0.9415 on unseen data, demonstrating very good generalization. During training and validation, the best performing model—saved at epoch 648—achieved a validation F1-score of 0.9409, indicating that the model performs even better on the test set, which is a strong sign of robustness and generalization rather than overfitting. The total training time to reach the optimal model was 322 minutes, shorter than two of the three benchmarked methods, demonstrating its suitability for real-time and resource-constrained applications.

\begin{figure}[h]
    \centering
    \includegraphics[width=0.9\columnwidth]{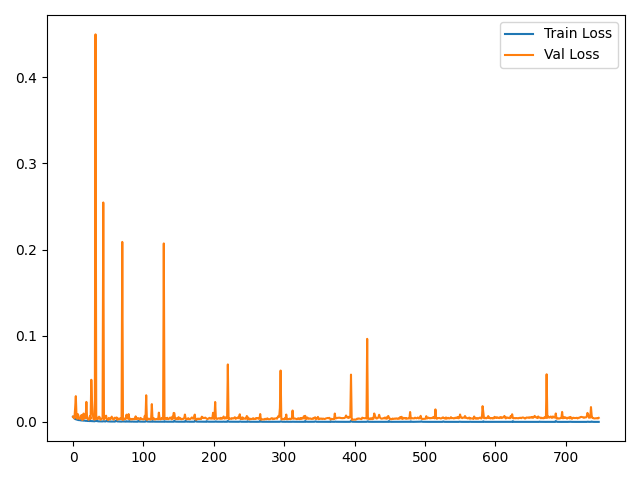}
    \caption{Training and validation loss graphs the proposed method.}
    \label{fig:ProposedMethod_Loss}
\end{figure}

Figure \ref{fig:ProposedMethod_Loss} shows the training and validation loss graphs over time. While some fluctuations and spikes are observed in the validation loss, the overall trend closely follows the training loss without upward pattern. This consistency suggests that overfitting did not occur, further supporting the model's reliability.

\begin{figure}[h]
    \centering
    \includegraphics[width=0.9\columnwidth]{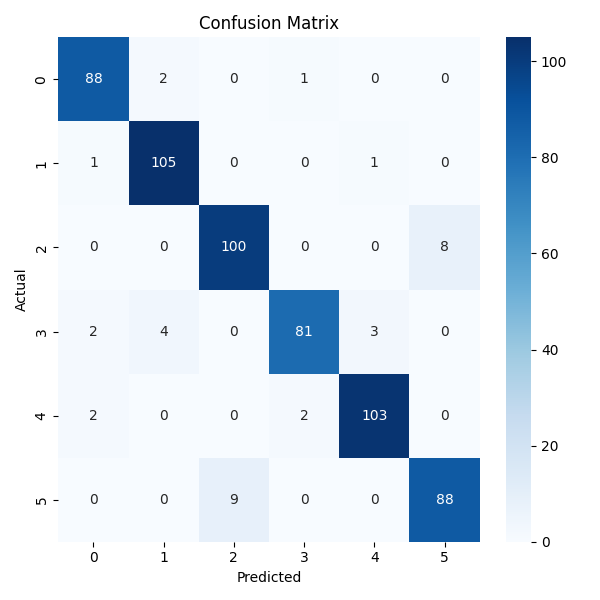}
    \caption{Training and validation loss graphs the proposed method.}
    \label{fig:ProposedMethod_Confusion}
\end{figure}

As shown in Figure \ref{fig:ProposedMethod_Confusion}, the confusion matrix further supports the model's strong performance across all activity classes. Most predictions are concentrated along the diagonal, indicating high true positive rates with only minor misclassifications. 

The classes shown in the matrix are indexed as {Cooking: 0, Drinking: 1, Eating: 2, Getting up: 3, Sitting down: 4, Washing up the dishes: 5}. As shown in the following sections, our approach achieves higher per-class accuracy, particularly for visually or temporally similar classes, highlighting the model’s ability to capture key spatiotemporal features.

Figures \ref{fig:ProposedMethod_Loss} and \ref{fig:ProposedMethod_Confusion} show high overall performance and good balance across class distributions, indicating effective generalization to unseen samples. The lightweight design and reduced parameter count contribute significantly to training efficiency.

\subsection{Comparison with Baseline Models}
In addition to evaluating the proposed model independently, its performance is also compared against three high-performing networks within the same class of 3D-CNNs. These baseline models were originally designed for three-channel RGB video input. Since the event-based input data in this research is structured similarly to single-channel grayscale frames, the input layers of these networks are modified to be compatible with one channel instead of three. Accordingly, the original pretrained weights—tuned for RGB input—could not be reused, due to their reliance on inter-channel correlations. Therefore, all benchmark models were retrained from scratch on our dataset using the same training pipeline to allow a fair comparison.

\subsubsection{Baseline Models} 
The three 3D-CNN models selected for benchmarking are: C3D, MC3\_18, and ResNet3D. The models are briefly introduced below, followed by a quantitative comparison of their performance.

\textbf{C3D:}
Convolutional 3D or C3D is developed by the researchers at Meta \cite{M1}. It is a widely used architecture for spatiotemporal feature learning in videos. It utilizes 3D convolutional and pooling operations to jointly capture spatial and temporal information across consecutive frames of videos. This enables more effective modeling of motion dynamics compared to traditional 2D ConvNets, which process frames independently and lack temporal awareness. C3D employs a homogeneous architecture with 3×3×3 convolution kernels throughout, which has been empirically shown to be effective for modeling dynamic visual patterns. Trained on large-scale video datasets such as Sports-1M, C3D provides compact, efficient, and transferable features that perform well across various video analysis tasks. Due to its simplicity and strong benchmark performance, it serves as a standard reference model for evaluating new spatiotemporal architectures, including the one proposed in this work.

\textbf{R3D:}
3D ResNet (R3D) \cite{M2}, extends the 2D residual learning framework to spatiotemporal data by incorporating 3D convolutions across all layers, enabling direct modeling of motion dynamics in video. In contrast to 2D CNNs that process individual frames, R3D learns spatiotemporal features jointly by convolving over both space and time. This architecture preserves the benefits of residual connections—such as improved optimization and gradient flow—while adapting them for volumetric video input. Although full 3D convolutions increase computational overhead, empirical studies show that R3D consistently outperforms 2D counterparts on large scale video benchmarks, validating the importance of explicitly modeling temporal structure for effective video understanding.

\textbf{MC3\_18:}
MC3-18 \cite{M3} is a hybrid architecture that combines elements of both 2D and 3D convolutional networks for video analysis. Based on the ResNet-18 \cite{M3} backbone, it integrates 3D convolutions in the early layers to capture short-term temporal dynamics, while employing 2D convolutions in the later layers to reduce computational cost. This selective use of spatiotemporal modeling results in a balanced trade-off between representational power and efficiency. Empirical studies demonstrate that MC3-18 achieves performance close to fully 3D architectures such as R3D, while using significantly fewer parameters and computations. On standard action recognition benchmarks, it outperforms 2D models and provides a competitive yet efficient baseline for spatiotemporal feature learning.

\subsubsection{Quantitative Comparison}
Table \ref{tab:3} presents the evaluation metrics for the baseline models and the proposed structure, as can be seen there our method performs better than all models.

\begin{table}[htbp]
\centering
\caption{Performance Comparison of Different Networks}
\setlength{\tabcolsep}{3pt} % default is 6pt
\renewcommand{\arraystretch}{1.1} % row height
\begin{tabular}{|l|c|c|c|c|}
\hline
\textbf{Network} & \textbf{F1-Score} & \textbf{Accuracy} & \textbf{Val Loss} & \textbf{Time (min)} \\
\hline
C3D & 0.6408 & 69.17\% & 0.0104 & \textbf{74} \\
\hline
ResNet3D & 0.9131 & 91.33\% & 0.4619 & 344 \\
\hline
MC3\_18 & 0.8622 & 86.67\% & 0.3388 & 948 \\
\hline
Proposed Method & \textbf{0.9415} & \textbf{94.17\%} & \textbf{0.0049} & 323 \\
\hline
\end{tabular}
\label{tab:3}
\end{table}

As shown in the table, the proposed method outperforms all other architectures across most metrics, with the exception of training time. In this case, C3D achieves the shortest training duration, followed by the proposed method. However, despite its fast training, C3D exhibits the weakest performance across all other evaluation metrics, indicating that low training time alone does not make it a viable option for this application.

\begin{figure*}[h]
    \centering
    
    \begin{subfigure}{0.33\textwidth}
        \centering
        \includegraphics[width=\linewidth]{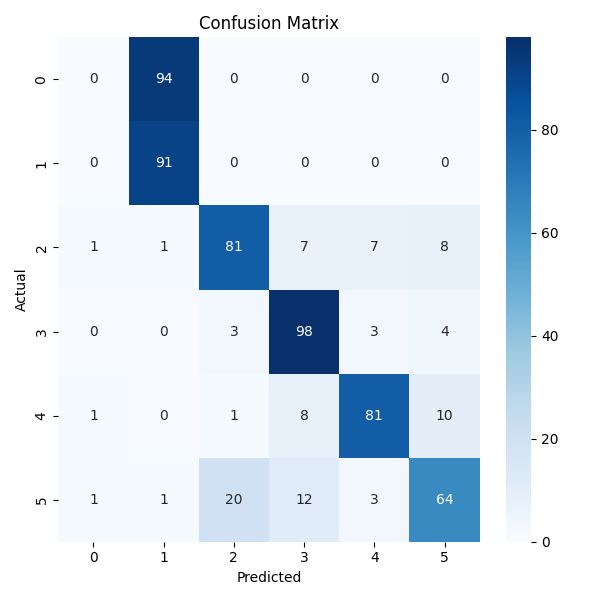}
        \caption{C3D}
    \end{subfigure}
    \hfill
    \begin{subfigure}{0.33\textwidth}
        \centering
        \includegraphics[width=\linewidth]{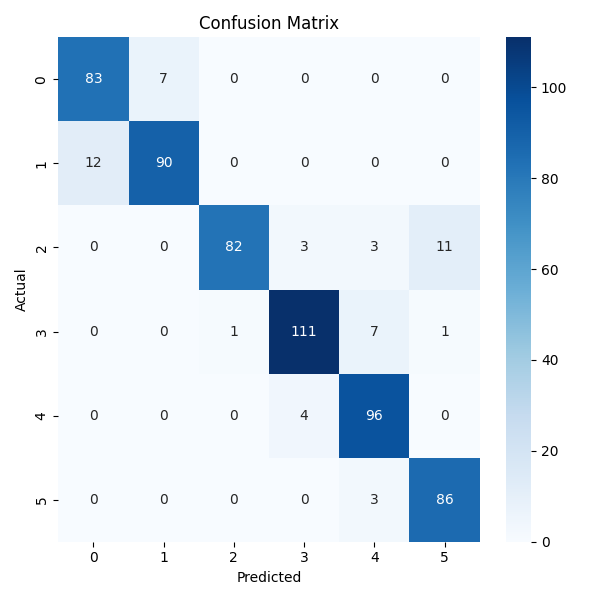}
        \caption{ResNet3D}
    \end{subfigure}
    \hfill
    \begin{subfigure}{0.33\textwidth}
        \centering
        \includegraphics[width=\linewidth]{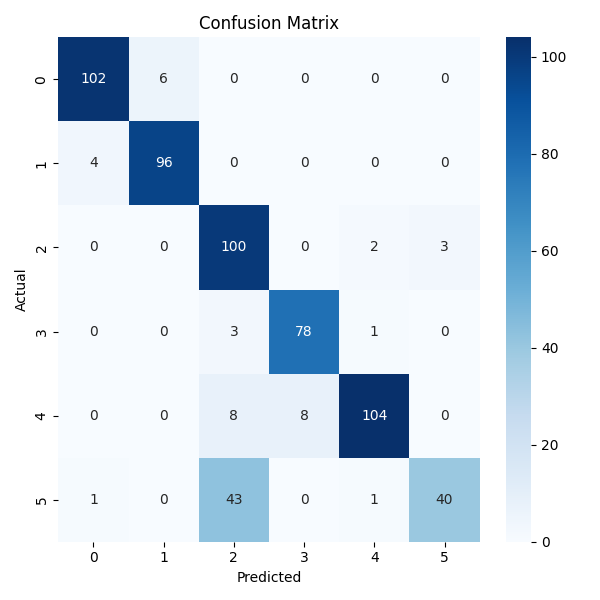}
        \caption{MC3\_18}
    \end{subfigure}
    
    \vskip\baselineskip % space between rows
    
    \begin{subfigure}{0.33\textwidth}
        \centering
        \includegraphics[width=\linewidth]{Figs/ProposedMethod_Confusion.png}
        \caption{Proposed method}
    \end{subfigure}
    
    \caption{Confusion matrices of networks: (a) C3D, (b) ResNet3D, (c) MC3\_18, (d) Proposed method.}
    \label{fig:ConfusionComparison}
\end{figure*}

Figure \ref{fig:ConfusionComparison} illustrates the confusion matrices for all methods. As shown, the proposed method exhibits the highest values along the main diagonal, indicating superior performance in correctly classifying the target classes compared to the other approaches.

\subsection{Qualitative Results}
\begin{figure*}[h]
    \centering

    \begin{subfigure}{0.95\textwidth}
        \centering
        \includegraphics[width=\linewidth]{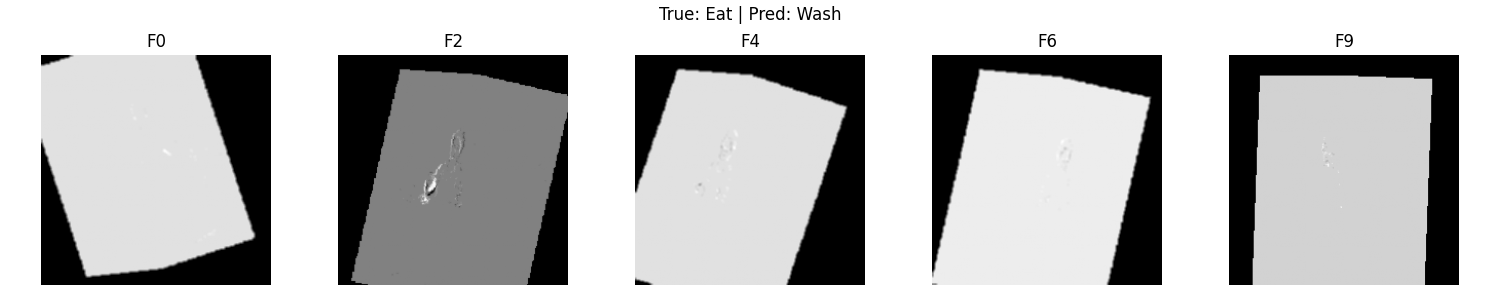}
        \caption{}
    \end{subfigure}

    \vspace{1ex}
    \begin{subfigure}{0.95\textwidth}
        \centering
        \includegraphics[width=\linewidth]{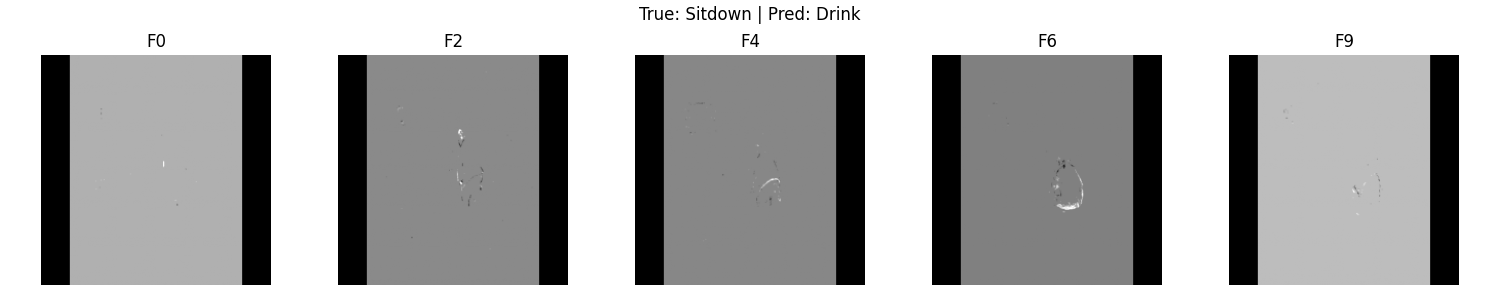}
        \caption{}
    \end{subfigure}

    \vspace{1ex}
    \begin{subfigure}{0.95\textwidth}
        \centering
        \includegraphics[width=\linewidth]{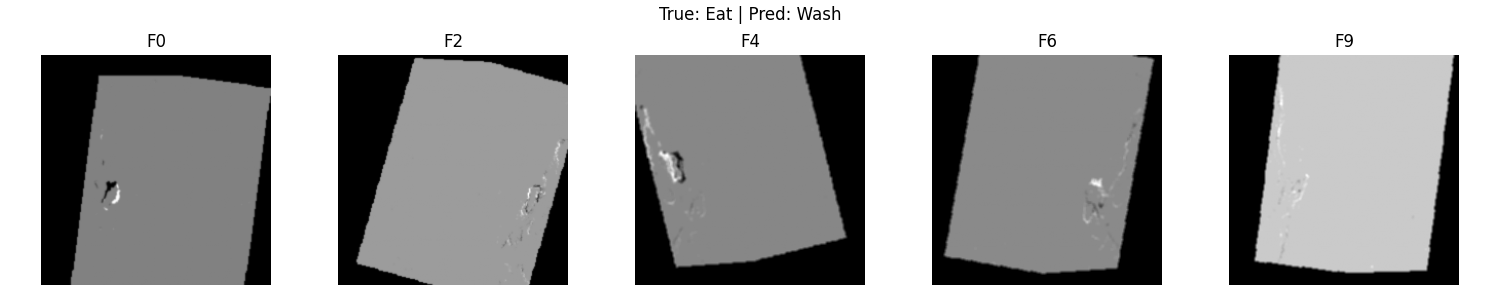}
        \caption{}
    \end{subfigure}

    \caption{Example frames from misclassified video samples classified by the proposed network.} 

    \label{fig:missclassified}
\end{figure*}

Figure \ref{fig:missclassified} shows three misclassified examples. For each video sample, five out of the ten frames are presented. As can be seen, even for a human observer it is not possible to identify the correct activity.

\subsection{Ablation Studies}
\subsubsection{Effect of Network Size}
A smaller version of the proposed network with half the channel size, leads to a 4\% drop in both accuracy and F1-score. Conversely, enlarging the network by doubling the channels size does not improve performance; instead, both accuracy and F1-score decrease by about 1\%, while training cost increases. These results indicate that the chosen network size achieves a good balance between capacity and efficiency and is well-suited to the dataset.

\subsubsection{Effect of Frame Rate}
The event stream was converted into 2D matrices and then sampled at a rate of 10 frames per each video. Reducing the rate to 5 frames per video decreased the accuracy to 89.33\% and the F1-score to 0.8939. Although the training time at this frame rate (78 minutes) was considerably shorter than the 323 minutes required at 10 frames, the nearly 5\% drop in accuracy made this trade-off undesirable. Increasing the rate to 20 resulted in a 2\% reduction in both accuracy and F1-score, while the training time increased to 332 minutes. These results suggest that 10 frames per video is sufficient to capture the necessary temporal context for most actions. The observed performance drop at higher frame rates may appear unexpected; however, it can be explained by the introduction of redundant information and noise, as well as the reduced generalization ability caused by incorporating more frames into the same network. Table \ref{tab:SizeComparison} summarizes the effect of varying the frame rate and network size on performance and training time. As shown in the table, the best accuracy is achieved by the proposed original architecture.

\renewcommand{\arraystretch}{1.3} % increases row height
\begin{table}[ht]
\centering
\caption{Performance comparison of different network configurations.}
\label{tab:SizeComparison}
\resizebox{\columnwidth}{!}{%
\begin{tabular}{|p{4cm}|c|c|c|c|}
\hline
\textbf{Network Configuration} & \textbf{F1-Score} & \textbf{Accuracy} & \textbf{Best Val Loss} & \textbf{Training Time} \\ \hline

Half Channels Size        & 0.9032 & 90.33\% & 0.0027 & 118 min \\ \hline
Double Channels Size      & 0.9332 & 93.33\% & \textbf{0.0024} & 231 min \\ \hline
Half Frames per Video             & 0.8939 & 89.33\% & 0.0031 & \textbf{79 min} \\ \hline
Double Frames per Video           & 0.9283 & 92.83\% & 0.0046 & 331 min \\ \hline
Original Network Architecture    & \textbf{0.9415} & \textbf{94.17\%} & 0.0049 & 323 min \\ \hline
\end{tabular}%
}
\end{table}

\subsubsection{Discussion of Strengths and Limitations}
\textbf{Strengths:}
\begin{itemize}
    \item The proposed compact 3D CNN architecture achieves fast inference while maintaining strong generalization across different datasets.
    \item Class imbalance is effectively mitigated through the use of focal loss and data augmentation strategies.
    \item Using event-based input data rather than conventional RGB frames offers potential privacy-preserving advantages.
\end{itemize}

\textbf{Limitations:}
\begin{itemize}
    \item Performance may degrade under extreme class imbalance.
    \item Subtle or overlapping motion patterns remain challenging to classify.
    \item Self-attention yields only marginal gains, and full transformer models were limited by computational resources.
\end{itemize}

\section{Conclusion} \label{Conclusion}
This work presented a lightweight 3D CNN architecture for human activity recognition using event-based vision data. By leveraging the inherent advantages of event cameras, the proposed method addresses key limitations of traditional frame-based HAR approaches, including privacy concerns. The network’s compact design, coupled with focal loss and targeted augmentation strategies, enables effective handling of class imbalance while maintaining high training efficiency which is an essential requirement for deployment on edge devices.

Experimental evaluations show that the proposed model outperforms widely adopted 3D-CNN baselines, including C3D, ResNet3D, and MC3\_18. With an F1-score of 0.91415 and an accuracy of 94.17\%, the model demonstrates strong generalization capability, particularly in distinguishing visually and temporally similar activities. These results confirm its effectiveness in extracting discriminative spatiotemporal features from low-resolution event-based inputs. In addition to its high accuracy, the model has a short training time.

Future work will investigate more advanced attention mechanisms, end-to-end event-stream processing without intermediate frame conversion—particularly through spiking neural network (SNN) approaches and adaptive temporal resolution strategies to further enhance recognition accuracy while preserving computational efficiency.

In summary, the proposed lightweight 3D-CNN provides an effective, privacy-preserving, and resource-efficient solution for real-time HAR, making it well suited for applications in healthcare, surveillance, and smart environments where both performance and privacy are paramount.

\bibliographystyle{IEEEtran}
\bibliography{bibliography}

% \begin{IEEEbiography}[{\includegraphics[width=1in,height=1.25in,clip,keepaspectratio]{Figures_Authors/mehdi.jpg}}]{Mehdi Sefidgar Dilmaghani} received his BSc, MSc, and PhD degrees in Electronics Engineering from the University of Tabriz in 2012, K. N. Toosi University of Technology in 2016, and the University of Galway in 2025, respectively. Prior to his PhD research, he focused on electronic implementation of signal processing algorithms and wavelets. During and after his PhD, his main research interests have been in computer vision, deep learning, and neuromorphic sensors (event cameras) for high-accuracy and privacy-preserving applications. 
% \end{IEEEbiography}

% \begin{IEEEbiography}[{\includegraphics[width=1in,height=1.25in,clip,keepaspectratio]{Figures_Authors/peter.png}}]
% {PETER CORCORAN} (Fellow, IEEE)  holds the Personal Chair in electronic engineering with the College of Science and Engineering, University of Galway. He was a Co-Founder of several start-up companies, notably FotoNation. He has over 600 technical publications and patents, over 100 peer-reviewed journal articles, 120 international conference papers; and a co-inventor of more than 300 granted U.S. patents. He is an IEEE Fellow, recognized for his contributions to digital camera technologies, notably in-camera red-eye correction and facial detection. He is a member of the IEEE Consumer Electronics Society, for over 25 years. He is the former Editors-in-Chief of IEEE Consumer Electronics Magazine.
% \end{IEEEbiography}

\EOD
\end{document}